\documentclass[a4paper,twoside]{article}

\usepackage{epsfig}
\usepackage{url}
\usepackage{calc}
\usepackage{hyperref}
\usepackage{multicol}
\usepackage{pslatex}
\usepackage{apalike}
\usepackage{algorithm2e}
\usepackage{natbib}
\usepackage[bottom]{footmisc}
\usepackage{SCITEPRESS}     

\begin{document}

\title{On the Effectiveness of Large Language Models in Automating Categorization of Scientific Texts}


\author{ 
\authorname{Gautam Kishore Shahi\orcidAuthor{0000-0001-6168-0132} and Oliver Hummel\orcidAuthor{0009-0007-3826-9477} }
\affiliation{University of Applied Sciences, Mannheim, Germany}
\email{\{g.shahi,o.hummel\}@hs-mannheim.de}
}

\abstract{The rapid advancement of Large Language Models (LLMs) has led to a multitude of application opportunities. One traditional task for Information Retrieval systems is the summarization and classification of texts, both of which are important for supporting humans in navigating large literature bodies as they e.g. exist with scientific publications. Due to this rapidly growing body of scientific knowledge, recent research has been aiming at building research information systems that not only offer traditional keyword search capabilities, but also novel features such as the automatic detection of research areas that are present at knowledge-intensive organizations in academia and industry. To facilitate this idea, we present the results obtained from evaluating a variety of LLMs in their ability to sort scientific publications into hierarchical classifications systems. Using the FORC dataset as ground truth data, we have found that recent LLMs (such as Meta's Llama 3.1) are able to reach an accuracy of up to 0.82, which is up to 0.08 better than traditional BERT models.}


\onecolumn \maketitle \normalsize \setcounter{footnote}{0} \vfill

\section{\uppercase{Introduction}}
\label{sec:introduction}

The amount of scholarly texts is consistently increasing; around 2.5 million research articles are published yearly \citep{rabby2024fine}. Due to this enormous increase, the classification of (scientific) texts has been attracting even  more attention in recent years \citep{bornmann2021growth}. 
Classifying the research area of scientific texts requires significant domain knowledge in various complex research fields. Hence, manual classification is challenging and time-consuming for librarians and limits the number of texts that can be classified manually \citep{zhang2023usage}. Moreover, due to complex hierarchical classification schemes and their existing variety, classification of publications is also an unbeloved activity for researchers. Prominent examples of classification schemes include the Open Research Knowledge Graph (ORKG) \citep{auer2019towards}, Microsoft Academic Graph \citep{wang2020microsoft}, the Semantic Scholar Academic Graph \citep{kinney2023semantic}, ACM computing classification system \citep{rous2012major}, Dewey Decimal Classification (DDC) \citep{scott1998dewey}, and the ACL Anthology \citep{bird2008acl}. Moreover, the coverage of these schemes is often subject-specific, for example, the well-known ACM classification is merely limited to computer science topics. As another example, ORKG currently has no in-depth classification for the top-level domain Arts and Humanities in its taxonomy.\footnote{https://huggingface.co/spaces/rabuahmad/forcI-taxonomy/blob/main/taxonomy.json}

Additional challenges with the existing systems in terms of scalability are highlighted by the following examples. First, consider ORKG, which was only recently created by volunteers who assigned tags to scientific texts and were merely able to classify a few thousand publications so far. Thus, an automated classification engine could significantly help to increase its coverage more quickly. Similarly, Microsoft Academic Graph (MAG) has only been relying on existing fields of study for scientific texts from the Microsoft Academic website \citep{herrmannova2016analysis}, but did not explicitly analyze or apply them. DDC, eventually, still has trouble dealing with new research topics and interdisciplinary fields \citep{wang2009extensive}. 

Thus, within organizations such as universities, research institutes, or even large companies where numerous researchers and other knowledge creators are working in multiple diverse domains, categorizing texts still requires considerable manual effort, making it challenging to deal with the huge volume of created texts and the contained knowledge. Consequently, there is a need for automated subject tagging systems to efficiently manage the steadily increasing volume of scientific texts and general knowledge contained in institutional repositories and comprehensive digital archives.

With the growth in generative artificial intelligence (GAI), especially, Large Language Models (LLMs) \citep{zhao2023survey}, a new opportunity to automate this tedious task has become tangible. LLMs are Artificial Intelligence (AI) systems that are specialized in generating human-like text for tasks such as summarization, translation, content creation, and even coding. LLMs have already been applied for several use cases, such as analyzing scientific documents \citep{giglou2024llms4synthesis}
, writing scientific reviews \citep{mahapatra2024artificial}, or information extraction \citep{pertsas2024annotated}.
LLMs can be configured by setting parameters such as the so-called quantization for reduced resource usage or their temperature, which controls the degree of creativity in an LLM's answer. LLMs are applied to their respective tasks by using so-called prompts, which are essentially textual commands describing the desired task at hand. The proper engineering of these prompts plays an important role in achieving the desired results with a model invocation \citep{gao2023prompt}.

\subsection{Research Goals}
In this study, we aim to better understand the benefits and quality currently achievable when using ``off-the-shelf'' LLMs for the classification of scientific texts and hence propose the following research question for our study. 

\noindent \textbf{RQ:} 
How can LLMs be effectively used to perform accurate tagging of research areas based on existing taxonomies?

To answer this question, we decided to utilize an existing classification as well as an existing dataset that has been recently published -- our experiments are based upon the ORKG taxonomy.\footnote{https://orkg.org/fields} and the Field of Research Classification (FoRC) Shared Task dataset \citep{abu2024forc}. The FORC dataset has been compiled by collecting manuscripts from ORKG and arXiv and was categorized into five top-level domains taken from ORKG, since the ORKG taxonomy provides a proven -- although not yet fully complete -- hierarchical structure for the classification of scientific texts from various domains. 

For our evaluation, we used a number of publicly available LLMs to evaluate their performance in terms of finding the (presumably) correct classification that the human volunteers have attributed to each publication from the dataset. The candidate LLMs each classified 59,344 scientific texts  based on their titles and abstracts with different temperatures by applying two types of prompts -- zero-shot and few-shot prompts. In a zero-shot prompt, an LLM is primed with limited information, namely merely by providing the task and the requirement to identify the research area, while in a few-shot prompt, we explained the task and also provided an example of a scientific paper together with an appropriate research area. A detailed description of our prompts is provided in section \ref{sec:2.1}. We applied these prompts with different contemporary LLMs, such as Gemma or Llama 3.1, and gauged the results with precision and recall, to finally calculate the accuracy for each model. Hence, the key contributions of this paper are as follows; it presents:

\begin{itemize}
    \item finding the research areas of scientific texts scientific documents as a novel application for LLMs
    \item an investigation of the influence of prompt engineering and parameter tuning in optimizing the results
    \item initial results on the performance of recent open-source LLMs for the classification for scientific texts. 
\end{itemize}

In the remainder of this paper, we discuss the state of the art in section~\ref{sec:related} and the proposed approach itself in section~\ref{sec:method}. After that, we discuss the implementation of the proposed approach in section~\ref{sec:implementation} and present results in section~\ref{sec:result}. Finally, we conclude our work and discuss future work in section~\ref{sec:future}.

\section{\uppercase{Related Work}}
\label{sec:related}


Document classification is one of the primary tasks for classifying scholarly research that is usually either performed by librarians or by subject experts, where both groups are faced with individual challenges: while the former are usually no subject experts, the latter are normally not trained for using document classification schemes. Multiple such schemes have been developed in recent decades to structure and classify the growing amount of scientific and subject-specific documents, for instance, the ACM computing classification system \citep{rous2012major}, the Dewey Decimal Classification (DDC) \citep{scott1998dewey}, or the ACL Anthology \citep{bird2008acl}. However, despite (or because of) this variety of existing taxonomies, manual subject tagging still remains challenging and especially time-consuming. For example, previous research reported that applying the Dewey Decimal Classification to a diverse dataset taken from the Library of Congress \citep{frank2004predicting} took librarians roughly five minutes per publication, as they were only able to assign DDC categories to 10.92 publications per hour \citep{wiggins2009acquisitions}. 

Until now, the automated classification of scientific articles into their respective research fields is -- despite decades of research -- still rather an emerging discipline \citep{desale2014research} than a proven practice that can be applied in libraries, universities, or the knowledge management in large corporations. In previous works, multiple approaches have been applied for this challenge, for instance, the work of \citep{wang2009extensive} used a supervised machine-learning approach for assigning DDC identifiers to documents collected from the Library of Congress \citep{wiggins2009acquisitions}. \citep{golub2020automatic} used six different machine learning algorithms to classify the documents from the Swedish library, where a Support Vector Machine gave the best results in terms of accuracy of classifiers. \citep{jiang2020improving} used BERT, an early transformer model for the identification of the research area on previously annotated data. However, up until today, the automation of deriving document classifications has mainly been a supervised learning task that requires specific training data and a thorough validation of results. 

In addition, additional challenges, such as the deep nesting of many classification taxonomies and data sparseness in certain classes need to be taken into account when implementing classification with ``traditional'' supervised learning. In general, this is especially challenging due to the need for a significant amount of labelled training data \citep{kalyan2023survey} that is still hard to find today. However, the recent generation of Large Language Models is pre-trained on extensive, unlabelled text data and hence is supposed to be more proficient in generating high-quality results in text classification without additional training or finetuning.

Thus, with the recent advancement of Large Language Models, LLMs have already been tested for several generic tasks in scholarly writing and provided promising results. In one recent study, ChatGPT has been used for automated classification of undergraduate courses in an e-learning system and improved overall performance in terms of accuracy significantly \citep{young2024chatgpt}. In another study, \citep{abburi2023generative} used an LLM for the automatic evaluation of scientific texts (in German language) written by students to assign a grade. \citep{pal2024ai} proposed an approach for using ChatGPT to develop an algorithm for plagiarism-free scientific writing. \citep{mosca2023distinguishing} built a data set to detect machine-generated scientific papers and compared results with other benchmark models. However, to our knowledge, so far, LLMs have not been tested for the identification of research areas of scientific texts, and hence our work provides a novel insight for the current performance of off-the-shelf LLMs in this area. In a recent publication, we have already proposed a search engine for indexing scientific documents enhanced with research areas, and demonstrated the practical usability of such subject tagging \citep{shahi2024enhancing}, e.g. in a search for domain experts. 


\section{\uppercase{Foundations}}
\label{sec:2.1}

The key aspect of GAI that separates it from other forms of artificial intelligence (AI), is that it is not primarily dedicated to analysing (numerical) data or acting based on such data like ``traditional'' AI (i.e. machine learning approaches) that has been used for this purpose in the past. Instead, GAI focuses on creating new content by using the data it was trained on \citep{hacker2023regulating, murphy2022probabilistic}. \textit{The term GAI thus refers to computational approaches which are capable of producing apparently new, meaningful content such as text, images, or audio} \citep{feuerriegel2024generative}.

Modern GAI for texts utilizes so-called Large Language Models that are trained on massive datasets to acquire different capabilities such as content generation or text summarization by learning statistical relationships of words \citep{wang2024gpt}. Modern LLMs are developed based on the so-called transformer architecture and trained on extensive corpora collected from public sources such as Web Crawls or GitHub using self- and human-supervised learning, enabling them to capture complex language patterns and contextual relationships \citep{perelkiewicz2024review}. Hence, LLMs can also be used for other quite diverse applications in natural language processing such as text summarization or data annotation. The well-known ChatGPT, launched by OpenAI, based on the Generative Pre-trained Transformer (GPT) architecture \citep{nah2023activity} is one such GAI that has been trained on a huge body (i.e., a significant part of the public WWW) of text.


Two important factors that can influence the perceived performance of LLMs for a certain task are prompt engineering and the so-called temperature used by the model.
Prompting strategies in LLMs include writing instructions for the models that are intended to guide responses effectively. Common techniques include providing context, step-by-step instructions, and examples to improve accuracy and relevance. A more detailed description of prompting strategies can be found in \citep{al2025evaluation}.  
In our experiments, we used the following two prompting strategies for evaluating the LLM and provide some examples fitting our context in Table \ref{prompt}.

\begin{itemize}
     \item \textbf{Zero-shot} In this approach, we ask LLMs to annotate the research area without providing any description or examples, which employs a simple and straightforward approach for extracting the research area. Zero-shot is also known as Vanilla Prompt, which does not take any prior knowledge or specific training on that task. It uses the pre-trained general ``intelligence'' of an LLM to obtain the research area for a scientific text. 
 \item  \textbf{Few-shot} In this case, learning is done based on context, where the model takes some description and an example for the research area as defined by ORKG taxonomy \citep{auer2019towards} to better understand its task. The model takes this input and provides answers based on the given information in conjunction with its general knowledge.
\end{itemize}

Moreover, we have employed the LLMs with varying temperatures, which adjusts the randomness of the responses given by an LLM. Lower temperatures give more focused and deterministic results, while higher temperatures generate more diverse and ``creative'' results. The value of the parameter starts from 0; however, we limited it to the range of [0,1] as temperatures above 1 result in a very high degree of randomness and neither coherence nor good reproducibility.

Different LLMs are trained in different objectives and with different training datasets, which is likely affecting their strengths in producing helpful results for our context. Hence, we used a set of temperatures and prompts for different LLMs, such as Llama and Gemma (more details follow in section \ref{llm}), to let them identify the research area of scientific texts from the test data set.

\begin{table*}[!htb]
\centering
\caption{Prompting strategies for determining research area from scientific texts}
\begin{tabular}{|p{7.5cm}|p{7.7cm}|}
\hline
\textbf{Zero Shot (Prompt 1)} & \textbf{Few Shot (Prompt 2)} \\
\hline
Suppose you are a data annotator who finds the research area of scientific texts. & Suppose you are a data annotator who finds the research area of scientific texts. \\

 & You are provided with scientific texts. Your task is to read texts and determine which research area from the list best represents the content of the scientific texts. Here is the hierarchy for each research: taxonomy of research field extracted from ORKG\footnote{https://orkg.org/fields})\\ 
 
Scientific text to annotate is \textit{factors influencing the behavioral intention to adopt a technological innovation from a developing country context: the case of mobile augmented reality games}  & scientific texts to annotate is \textit{comparative analysis of algorithms for identifying amplifications and deletions in array cgh data}

\\ 

 & \\
 
Assign a research area to the given scientific texts and provide it as output & Assign a research area from the given taxonomy above and provide it as output\\
  & \\
 \textbf{\textit{Expected Output}}: [Social and Behavioral Sciences] \textbf{\textit{Output}:} [Social Science]  
 
 &  \textbf{\textit{Expected Output}}: [Physical Sciences \& Mathematics] \textbf{\textit{Output}:} [Physical Sciences \& Mathematics]\\ \hline

\end{tabular}
\label{prompt}

\end{table*}

\section{\uppercase{Approach}}
\label{sec:method}
\begin{figure*}[!htbp]
  \centering
\includegraphics[width=0.6\textwidth]{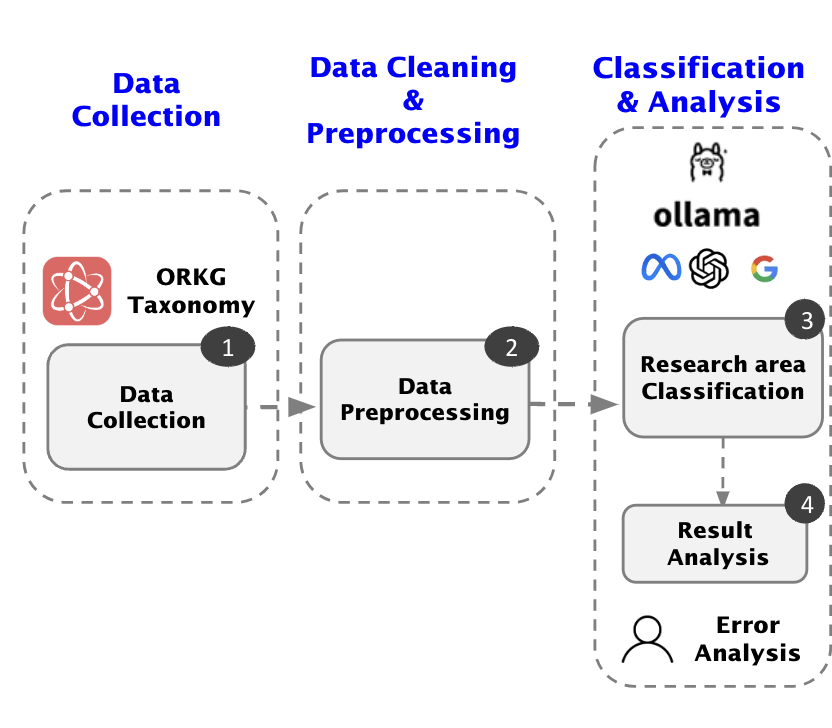}
  \caption{Methodology used in the identification of research area}
  \label{fig:method}
 \end{figure*}

To address the research question in our present study, we propose a methodology comprising data collection, data cleaning, and preprocessing, followed by the application of prompt engineering to classify the research areas of scientific texts. 
This process is summarized in Figure \ref{fig:method} and explained in more detail in the following subsections. 

Our approach is divided into three main parts, i.e., data collection, data cleaning, preprocessing, and classification and analysis. The first step involves the collection of the required dataset suitable for the study; in this case, we analysed scientific texts extracted from the FORC dataset; a detailed description of the data is provided in section \ref{data}. The second step involves data cleaning and preprocessing, which includes removing unwanted information such as formatting information before the texts were fed to the LLMs. Finally, the third and most crucial step involves the application of LLMs to identify the research area and analyze its results. At present, we aim at the prediction of the top-level domain from the ORKG classification as explained in Section \ref{data}. 


We employed four contemporary LLMs (cf. Table \ref{tab:llm}) with a small and medium amount of parameters ranging from 3.82b to 70.4b to classify the research areas of the selected texts. Each LLM was obtained from and executed with Ollama\footnote{https://ollama.com/library}.
To assess the performance of LLMs, we compared their results with those of traditional Bidirectional Encoder Representations from Transformers (BERT) models \citep{devlin2018bert}. A detailed explanation of the experimental setup is provided in Section \ref{sec:implementation}, while the results are discussed in Section \ref{sec:result}.


\begin{table*}[!htbp]
\centering
\caption{A short description of LLM models used in the study}
\begin{tabular}{|p{2.2cm}|p{5cm}|p{3cm}|p{1.8cm}|}
\hline
\textbf{Model)} & \textbf{Description}  & \textbf{No. of Parameters}  & \textbf{Release Date}\\
\hline

Gemma 2\footnote{https://blog.google/technology/developers/google-gemma-2/} & Gemma2 is a lightweight, state-of-the-art open models &  parameters-27.2B \& quantization-Q4\_0 & June 2024  \\ \hline
Llama 3.1\footnote{https://ai.meta.com/blog/meta-llama-3-1/} & Llama 3.1 70B is a multilingual model that has a significantly longer context length of 128K, state-of-the-art tool use, and overall stronger reasoning capabilities  & parameters-70.4B and quantization-Q4\_0  & July 2024  \\ \hline

Mistral Nemo\footnote{https://mistral.ai/news/mistral-nemo/} & Mistral NeMo offers a large context window of up to 128k tokens. Its reasoning, world knowledge, and coding accuracy are state-of-the-art in its size category & parameters-12.2B \& quantization-Q4\_0 & July 2024 \\ \hline
Phi\footnote{https://huggingface.co/microsoft/Phi-3.5-mini-instruct} &  Phi 3.5 is a lightweight, state-of-the-art open model built upon synthetic datasets & parameters-3.82B \& quantization-Q4\_0 & August 2024 \\ \hline

\end{tabular}
\label{tab:llm}

\end{table*}

\section{\uppercase{Experiments}}
\label{sec:implementation}

For our experiments, we have been using Ollama frameworks \citep{ollama}, an open source application that allows the easy running of LLMs on local hardware. Ollama provides an easy opportunity to run the model locally with a simple command-line interface that directly interacts with the LLMs and allows easy installation and implementation. Ollama allows downloading models with a given number of parameters. Currently, there are more than 3,100 models registered on Ollama (accessed on 6th August 2024)\footnote{https://ollama.com/search} by numerous different users. 

For this experiment, we utilized LLaMa (70 billion parameters), Mistral Nemo (12 billion parameters), Gemma (27 billion parameters), and Phi (3.8 billion parameters) as our foundational models. The respective model sizes are 2.4 GB for Phi, 15 GB for Gemma, 4.1 GB for Mistral Nemo, and 39 GB for LLaMa.  
The computational infrastructure consists of an in-house server equipped with four NVIDIA RTX A6000 GPUs, each with 48 GB of VRAM, 512 GB of system memory, and 8 TB of storage, ensuring sufficient resources for efficient model execution and experimentation.


We developed a Python program utilizing Ollama and LangChain, an open-source framework for building large language model (LLM) applications, to execute the LLM models for research area annotation. The generated results were systematically stored for evaluation. For LLM optimization, key parameters such as temperature and prompt strategies play a crucial role. To evaluate different configurations, we implemented the models with various combinations of temperature settings and prompts. In the zero-shot setting, only the temperature and task description were provided, whereas the few-shot setting involved different prompt combinations to refine the outputs. An example of zero-shot and few-shot prompt strategies is presented in Table \ref{prompt}.

\subsection{\uppercase{Dataset}}
\label{dataset}
We used the scientific texts collected by the FORC shared task \citep{abu2024forc}, which is based upon the ORKG taxonomy. FORC consists of scientific texts, mainly research papers with DOI, research area, abstract, title, and author information. The FORC initiative compiled scientific texts from open-source resources such as ORKG (CC0 1.0 Universal) and arXiv (CC0 1.0), whereas scientific text with non-English titles or abstracts were excluded. Each scientific text has been assigned a field of research based on ORKG taxonomy.\footnote{https://orkg.org/fields}. 

The ORKG taxonomy\footnote{\url{https://orkg.org/fields}} provides a structured framework for the systematic classification and exploration of research domains. This taxonomy is organized into five primary domains: \textit{Arts and Humanities, Engineering, Life Sciences, Physical Sciences and Mathematics, and Social and Behavioral Sciences.} Each of these top-level domains is hierarchically structured into two additional levels: subdomains and subjects. At the first sub-level, each primary domain is subdivided into specific research areas, which are further refined into specialized subjects. For instance, within the \textit{Physical Sciences and Mathematics} domain, the \textit{Computer Science} subdomain includes \textit{Artificial Intelligence} as a subject. In this study, for the time being, we consider the top-level domains to maintain a high-level perspective on the classification of research fields.

Overall, FORC provides a collection of 59,344 scientific texts, each categorized using a taxonomy of 123 Fields of Research (FoR). These are organized across three hierarchical levels and grouped into five top-level categories: \textit{Physical Sciences and Mathematics, Engineering, Life Sciences, Social and Behavioral Sciences, Arts and Humanities}. For each scientific text, we filtered only meaningful information useful for us, as described in table \ref{dataset}. We used DOI as a unique identifier and title with abstract for our classification model to predict the research area. 

\begin{table*}[!htb]
\centering
\caption{Description of  collected dataset}
\begin{tabular}{|p{2.5cm}|p{5cm}|p{4.5cm}|}
\hline
\textbf{Field} & \textbf{Description} & \textbf{Example} \\
\hline

DOI & A DOI (Digital Object Identifier) is a standardized unique number given to paper, and we used it as a unique identifier of paper & 10.1145/2736277.2741136 \\ \hline
Title & Title describes the title of the paper &  Crowd Fraud Detection in Internet Advertising \\ \hline
Abstract & Abstract of the paper describing the summary of paper & "the rise of crowdsourcing brings new types of malpractices in internet advertising. one can easily hire web workers through malicious crowdsourcing platforms to attack other advertisers.... \\ \hline
Research area & research area defined based on the ORKG taxonomy, and it is a dependent variable for our prediction model & Engineering \\ \hline 
 
\end{tabular}
\label{data}

\end{table*}

\subsubsection{Data Cleaning and Preprocessing}
After collecting the dataset, we cleaned and preprocessed the data to remove unwanted information from the scientific text. This section describes the steps involved in the data preprocessing and analysis. LLMs work as a black-box algorithm \citep{liu2024language}, and we do not have an internal functional model to provide the research area of scientific texts, so we provided preprocessed data for all models to maintain fairness. After collecting the dataset, we removed unwanted information, such as URLs mentioned in the text, special characters in abstracts, and authors of the publication. We used title and abstract from all 59,344 scientific texts tagged by FORC to identify the research area and derived the accuracy to evaluate our LLM-based prediction model.

\subsection{Evaluated LLMs}
\label{llm}
We employed four open-source LLMs for our classification experiments, namely, Llama (Meta), Gemma (Google), Nemo (Mistral), and Phi (Microsoft). A detailed description of each selected LLM is given below, together with important metadata, which is provided in Table \ref{tab:llm}. After that, the experimental setup is explained in section \ref{sec:implementation}.  
\begin{itemize}
\item \textbf{Gemma} We used the recent version Gemma 2 \citep{team2024gemma}, which is a family of lightweight, state-of-the-art open-source models that are advertised as high-performing and efficient models by Google. They are currently available in two sizes; we have used Gemma 2 with 27 million parameters. Gemma was trained on web documents and using mathematics, outperforming other models in 11 of 18 text-based tasks in terms of efficiency \citep{team2024gemma}. 

\item \textbf{Llama} We used the latest version, which was Llama 3.1 at the time of writing. Llama is developed and released by Meta \citep{touvron2023Llama}; there are currently three versions of Llama with different sizes of 8b, 70b, and 405b parameters; we have used Llama 3.1 with 70b parameters. Llama is trained on publicly available data without resorting to proprietary datasets. For the training, different data sources, such as CommonCrawl and GitHub, were used.

\item \textbf{Mistral Nemo} is the latest LLM developed jointly by Mistral AI and NVIDIA AI with 12B parameters and a context window of up to 128k tokens. Mistral Nemo outperformed the prior Mistral model LLama 3 and Gemma 2 in terms of efficiency and effectiveness despite having fewer parameters. \footnote{\url{https://mistral.ai/news/mistral-nemo/}}

\item \textbf{Phi} is a family of lightweight, open large language models developed by Microsoft that are designed to be efficient and accessible. The "Phi-3" family includes models with 3 billion (3B) and 14 billion (14B) parameters, classified as "Mini" and "Medium" respectively. Phi outperforms\footnote{https://azure.microsoft.com/de-de/blog/new-models-added-to-the-phi-3-family-available-on-microsoft-azure/} Gemini 1.0 Pro, and the model is trained on high-quality educational data, newly created synthetic, “textbook-like” data, which should make it especially suitable for use for classification tasks of the scientific domain.
\end{itemize}

\subsection{Baseline models}
Given our collected corpus of scientific texts, we chose the following two classification models as state state-of-the-art methods for result comparison:
\textbf{BERT} \citep{devlin2018bert} is a widely used pre-trained model for text classification. The model has been applied to various classification tasks and evaluated across multiple domains, including the classification of text related to COVID-19 \citep{shahifakecovid}. BERT utilizes a bidirectional transformer mechanism, allowing it to capture contextual relationships in text more effectively than traditional models. It has demonstrated state-of-the-art performance in numerous natural language processing (NLP) benchmarks, making it a strong candidate for research area classification.  \\
\textbf{BiLSTM} (Bidirectional Long Short-Term Memory) \citep{huang2015bidirectional} is a recurrent neural network (RNN) designed for text classification, capturing input flows in both forward and backward directions. It has been successfully applied to various NLP tasks, including the classification of scientific texts \citep{enamoto2022multi}. BiLSTM enhances sequential data processing by preserving long-range dependencies, reducing the vanishing gradient problem, and improving contextual understanding. Its ability to capture bidirectional dependencies makes it effective in tasks requiring nuanced text comprehension.

For the implementation of the baseline models, we developed a Python program and retrieved the pre-trained models from Hugging Face.\footnote{\url{https://huggingface.co/}} Both BERT and BiLSTM were implemented using the models available on this platform.

\section{\uppercase{Results}}
\label{sec:result}

We evaluated the baseline models and the selected LLMs according to the two explained prompts and temperature values, from 0.2 up to 1.0 with all nearly 60.000 titles and abstracts from the FORC dataset. Depending on the size of the model, this took about 3 hours (for Phi) and about 22 hours (for Llama) with the others in between. The accuracy obtained by the state of the art baseline models is shown in Table \ref{result:ml}, the accuracy for the LLMs in Table \ref{result:llm1} and \ref{result:llm2} respectively. 

Overall, the Llama model achieved the best results, and few-shot prompt strategies outperformed the models with zero-shot prompts. Also, increasing the temperature seems to help in obtaining better results; however, after reaching the best performance at 0.8, quality starts decreasing with the one larger temperature value we have tested. 

\begin{table}[!htb]
\centering
\caption{Accuracy of State of Arts models.}
\begin{tabular}{|p{2cm}|c|}
\hline
\textbf{Model} & \textbf{Accuracy} \\
\hline
BERT & 0.74  \\\hline
BiLSTM & 0.66  \\\hline

\end{tabular}
\label{result:ml}
\end{table}

\begin{table}[!htb]
\centering
\caption{Accuracy of LLMs according to prompt 1 (Zero-shot) with different temperatures.}
\begin{tabular}{|p{2cm}|c|c|c|c|c|}
\hline
\textbf{Parameter} & \textbf{0.2} & \textbf{0.4} & \textbf{0.6} & \textbf{0.8} & \textbf{1} \\
\textbf{Model} & & & & & \\
\hline
Gemma & 0.18 & 0.24 & 0.30 & 0.42 & 0.50 \\\hline
Llama & 0.24 & 0.28 & 0.48 & \textbf{0.62} & 0.60 \\\hline
Mistral Nemo & 0.10 & 0.18 & 0.34 & 0.52 & 0.44 \\\hline
Phi & 0.30 & 0.18 & 0.24 & 0.48 & 0.42 \\\hline
\end{tabular}
\label{result:llm1}
\end{table}

\begin{table}[!htb]
\centering
\caption{Accuracy of different LLMs according to prompt 2 (Few-shot) with different temperatures.}
\begin{tabular}{|p{2cm}|c|c|c|c|c|}
\hline
\textbf{Parameter} & \textbf{0.2} & \textbf{0.4} & \textbf{0.6} & \textbf{0.8} & \textbf{1} \\
\textbf{Model} & & & & & \\
\hline
Gemma & 0.18 & 0.24 & 0.40 & 0.66 & 0.62 \\\hline
Llama & 0.34 & 0.38 & 0.64 & \textbf{0.82} & 0.72 \\\hline
Mistral Nemo & 0.14 & 0.38 & 0.44 & 0.76 & 0.62 \\\hline
Phi & 0.30 & 0.18 & 0.44 & 0.58 & 0.62 \\\hline
\end{tabular}
\label{result:llm2}
\end{table}

As is visible in the tables, the relatively high temperature of 0.8 seems to work best with most models, sometimes 1.0 provides even better results. Not surprisingly, the model largest in terms of parameters, i.e. Llama with 70 billion parameters, delivered the best results in our experiments and achieved an Accuracy of 0.82 for the few-shot prompt.

\subsection{Error Analysis}


In addition to the automated evaluation presented earlier, we conducted a manual error analysis to assess the performance of the best-performing model. Specifically, we focused on analyzing the misclassified instances produced by LLaMa 3.1 to gain deeper insights into its errors.
To achieve this, we randomly sampled 100 scientific texts from the incorrectly classified results and manually analyzed them. This manual review provided qualitative insights into common causes of misclassification (such as short abstracts, missing abstracts, or missing titles), helping to identify potential reasons for incorrect predictions. The findings support the automated evaluation, indicating that LLaMa 3.1 produces reliable and usable results.

\section{\uppercase{Summary}}

In the present study, we have tested a number of state-of-the-art open-source LLMs for the identification of research areas of scientific documents based on 59,344 abstracts taken from the FORC dataset \citep{abu2024forc}. Overall, the LLMs are capable of identifying the research areas pretty well, as underlined by accuracy of up to 0.82. However, for the time being, our study merely tackled the highest level of the FORC classification, leaving a lot of room for future work aiming on the lower levels as well.
 
In general, automatic tagging of scientific (and other) texts is still an ongoing challenge that will require future work, as there still exists a lack of cross-domain datasets that cover common subject areas. Hence, another limitation of our study is that the dataset we have used was taken from a previous work with a different goal, so that it was not a perfect fit for the task at hand in terms of overall coverage. E.g., we found that some of the classes are currently not well covered in the dataset, as there are currently no sub-classes for Arts and Humanities, which would make it hard to generalize our topic classification results for general libraries that have to deal with texts from virtually all knowledge areas. 

Several other taxonomy systems exist for tagging scientific texts, such as the ACM classification system or the Dewey Decimal Classification (DDC). However, they still suffer from similar limitations, such as non-existent gold standards. Consequently, evaluating LLMs for these taxonomies must be covered in future work that is also likely to feed back interesting improvement potential for the taxonomies.

Although large language models have exhibited remarkable efficacy in addressing a wide range of challenges, their deployment for classification tasks remains fraught with significant challenges. Specifically, LLMs require substantial computational resources, including high-performance GPUs and extensive memory capacities, resulting in considerable economic implications, which is another indicator for the substantial efforts that are required in the near future to achieve a solid understanding of how well LLMs can be used for various tasks.


\section{\uppercase{Conclusion \& Future Work}}
\label{sec:future}

This study systematically investigates the application of large language models (LLMs) for automated research area classification in scientific literature. The proposed methodology was implemented on the FORC dataset, taken from the ORKG initiative, employing two distinct prompt engineering strategies while optimizing the temperature parameter to enhance classification performance. The study used the five top-level domains of ORKG taxonomies to classify research domains, which can obviously be extended to predict subdomains and subjects of scientific text in the near future. Four LLMs, namely Gemma, Llama, Nemo and Phi were rigorously evaluated against two baseline BERT models using a large dataset of almost 60,000 publications. Results are indicating that modern LLMs are superior to previous models and that few-shot prompting significantly improves classification accuracy. Among the models tested, Llama achieved the highest accuracy, making it the most effective for research area identification.

Future research directions include leveraging additional state-of-the-art LLMs, addressing more fine-granular taxonomy levels, and integrating alternative classification schemes, such as the ACM Computing Classification System and the Dewey Decimal Classification (DDC), to refine scientific text classification even further. The proposed method can also be deployed in institutional research centers and academic libraries to systematically identify and categorize forthcoming scholarly publications, enhancing knowledge organization and retrieval in academic and industrial research environments.

\section*{Data Sharing} 
We have conducted all experiments on a macro level following
strict data access, storage, and auditing procedures for the sake of
accountability. We release the processed data used in the study along with minimal code to replicate the model for the community. The code and the dataset are available at GitHub here.\footnote{\url{https://github.com/Gautamshahi/LLM4ResearchArea}}



\section*{\uppercase{Acknowledgements}} 
The work has been carried out under the TransforMA project. Authors disclosed receipt of the following financial support for the research, authorship, and/or publication of this article. This project has received funding from the federal-state initiative "Innovative Hochschule" of the Federal Ministry of Education and Research (BMBF) in Germany. 


\bibliographystyle{apalike}
{\small
\bibliography{example}}


\end{document}